\documentclass[letterpaper]{article} 
\usepackage{aaai2026}  
\usepackage{times}  
\usepackage{helvet}  
\usepackage{courier}  
\usepackage[hyphens]{url}  
\usepackage{graphicx} 
\usepackage{natbib}  
\usepackage{caption} 
\usepackage{algorithm}
\usepackage{algorithmic}
\usepackage{amsmath}
\usepackage{multirow}
\usepackage{longtable}
\usepackage{graphicx}
\usepackage{float}
\usepackage{subfigure}
\usepackage{array}
\usepackage{pgfplots}
\usepackage{svg}
\usepackage{booktabs}
\usepackage{tabularx}
\usepackage{listings}
\usepackage{adjustbox}
\usepackage{caption}

\usepackage[table]{xcolor}

\usepackage{newfloat}
\usepackage{listings}
\DeclareCaptionStyle{ruled}{labelfont=normalfont,labelsep=colon,strut=off} 
\floatstyle{ruled}
\newfloat{listing}{tb}{lst}{}
\floatname{listing}{Listing}
\title{Start Small, Think Big: Curriculum-based Relative Policy Optimization for \\ Visual Grounding}

\author{
    Qingyang Yan\textsuperscript{\rm 1}\equalcontrib, 
    Guangyao Chen\textsuperscript{\rm 2}\equalcontrib\footnotemark[2], 
    Yixiong Zou\textsuperscript{\rm 1}\thanks{Corresponding author.}
}
\affiliations{
    \textsuperscript{\rm 1}School of Computer Science and Technology, Huazhong University of Science and Technology\\

    \textsuperscript{\rm 2}National Key Laboratory for Multimedia Information Processing, School of Computer Science, Peking University
}

\begin{document}

\maketitle

\begin{abstract}
Chain-of-Thought (CoT) prompting has recently shown significant promise across various NLP and computer vision tasks by explicitly generating intermediate reasoning steps. However, we find that reinforcement learning (RL)-based fine-tuned CoT reasoning can paradoxically degrade performance in Visual Grounding tasks, particularly as CoT outputs become lengthy or complex. Additionally, our analysis reveals that increased dataset size does not always enhance performance due to varying data complexities. Motivated by these findings, we propose \textbf{Curriculum-based Relative Policy Optimization (CuRPO)}, a novel training strategy that leverages CoT length and generalized Intersection over Union (gIoU) rewards as complexity indicators to progressively structure training data from simpler to more challenging examples. Extensive experiments on RefCOCO, RefCOCO+, RefCOCOg, and LISA datasets demonstrate the effectiveness of our approach. \textbf{CuRPO} consistently outperforms existing methods, including Visual-RFT, with notable improvements of up to +12.52 mAP on RefCOCO. Moreover, \textbf{CuRPO} exhibits exceptional efficiency and robustness, delivering strong localization performance even in few-shot learning scenarios, particularly benefiting tasks characterized by ambiguous and intricate textual descriptions.The code is released on \url{https://github.com/qyoung-yan/CuRPO}.
\end{abstract}
\section{Introduction}

Chain-of-Thought (CoT) prompting has recently garnered significant attention within both natural language processing (NLP) and computer vision research communities due to its capability to enhance model interpretability and performance by explicitly generating intermediate reasoning steps~\cite{wei2022chain, kojima2022large, ge2023chain}. These explicit reasoning processes have been successfully integrated with reinforcement learning techniques such as Group Relative Policy Optimization (GRPO)~\cite{liu2025visual}, significantly boosting performance across a variety of tasks including question answering, arithmetic reasoning, and visual reasoning tasks~\cite{chen2024visual, liu2025visual, tan2025reason}. Despite these successes, we observe a somewhat counterintuitive phenomenon in the domain of Visual Grounding: explicitly generating CoT reasoning steps does not consistently lead to improved performance and, in certain cases, may even degrade model accuracy, particularly when CoTs become overly lengthy or complex.

\begin{figure}[!tbp]
\centering
\includegraphics[width=\linewidth]{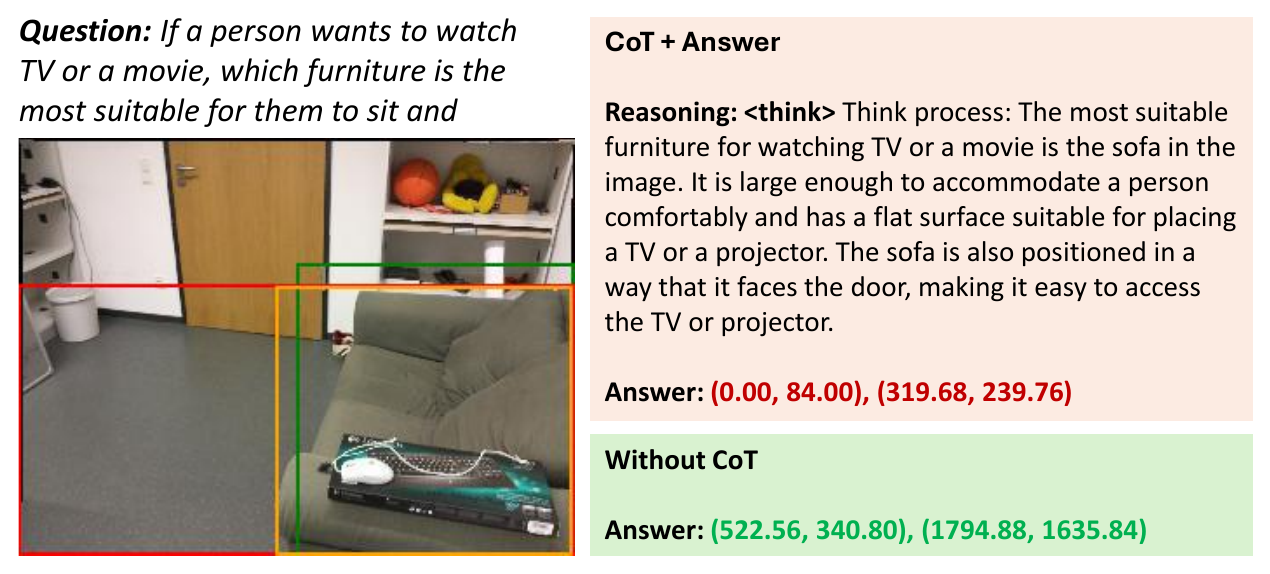}
\caption{Comparison of visual grounding results with and without CoT. The CoT-guided model produces an incorrect bounding box (red) due to misinterpretation of the textual context. In contrast, the model without CoT successfully identifies the correct furniture for watching TV or a movie (green bounding box).}
\label{fig:without_cot_better}
\end{figure}

As illustrated in Figure~\ref{fig:without_cot_better}, our experiments reveal that a GRPO-trained model generates incorrect bounding box predictions when explicitly prompted to produce CoT reasoning, while the same model successfully identifies the correct object without generating explicit CoT. This suggests that, within the context of Visual Grounding, explicitly generating CoT may introduce unnecessary complexity, thereby reducing localization accuracy. Further investigation into this phenomenon uncovered another crucial issue: increasing the size of the training dataset for Visual Grounding does not consistently result in improved model performance. Surprisingly, model performance can even deteriorate as more data is added, indicating the presence of varying levels of difficulty within larger datasets. This observation prompts us to reconsider conventional assumptions about data quantity and complexity, motivating a deeper exploration of whether and how the complexity and ordering of training examples might impact the learning dynamics and overall performance of visual grounding models.

Our proposed method, termed \textbf{Curriculum-based Relative Policy Optimization (CuRPO)}, explicitly leverages CoT length and reward indicators to structure training in a curriculum manner, progressively increasing task complexity. Specifically, \textbf{CuRPO} begins by generating multiple CoT-based reasoning outputs for each training sample, calculating their average length to quantify the inherent complexity of each example. Concurrently, we employ visual reward as an additional reward-based complexity measure, ensuring comprehensive evaluation of each data point’s difficulty. By sorting and grouping training examples according to these indicators, \textbf{CuRPO} systematically exposes the model first to simpler examples, gradually advancing to more challenging scenarios. This carefully structured progression allows the model to incrementally acquire robust reasoning and precise localization capabilities, significantly enhancing its overall accuracy and training stability, even in settings with limited training data.

Extensive experiments across multiple visual grounding benchmarks, including RefCOCO, RefCOCO+, RefCOCOg, and LISA, substantiate the effectiveness of our proposed method. \textbf{CuRPO} consistently demonstrates substantial performance improvements over existing baselines, notably outperforming Visual-RFT~\cite{liu2025visual} by large margins, achieving up to +12.52 mAP on the RefCOCO dataset. Importantly, our approach exhibits remarkable efficiency and effectiveness under few-shot learning scenarios, delivering strong localization performance even with minimal training samples and showing their efficacy in tackling datasets characterized by ambiguous and complex textual descriptions.

In summary, our contributions are threefold:
\begin{itemize}
\item We identify and empirically verify that explicitly generating CoT reasoning may negatively impact visual grounding performance.
\item We introduce CoT length and reward-based sorting as novel indicators of data complexity, forming the basis for an effective curriculum training strategy (\textbf{CuRPO}).
\item Our \textbf{CuRPO} method demonstrates consistent performance gains across multiple visual grounding benchmarks, significantly improving accuracy, training stability, and exhibiting remarkable effectiveness in few-shot settings.
\end{itemize}

\section{Related Work}

\paragraph{Chain-of-Thought Reasoning.}
Chain-of-Thought (CoT) prompting asks language models to generate explicit intermediate steps before answering \cite{wei2022chain}.  By decomposing problems into smaller sub‑tasks, CoT improves arithmetic, commonsense, and code‑generation benchmarks \cite{zhang2024examination,ke2025survey}.  These gains, however, depend strongly on reasoning length and token budget.  Longer traces help on difficult items \cite{fu2022complexity,jin2024impact} but can degrade accuracy or efficiency on simpler ones due to “over‑thinking” \cite{nayab2024concise,chen2024more,chen2024not,Zou2024AttentionTM}.  Excessive length also hurts generalization when such sequences are under‑represented in pre‑training corpora \cite{exploring}.  
To mitigate these issues, \citet{wu2025more} propose a length‑adaptive scheduler that selects a “balanced’’ CoT depth based on task hardness and model size, recovering efficiency without sacrificing accuracy.  Overall, the evidence suggests that CoT is beneficial only when reasoning depth is matched to instance difficulty. Enumeration‑based search such as Self‑Consistency \cite{wangself} and Tree‑of‑Thought (ToT) sampling \cite{yao2023tree} improves robustness by exploring multiple reasoning paths and selecting the highest‑scoring answer. Recently, Chain‑of‑Preference Optimization (CPO) aligns each intermediate step with high‑quality ToT traces during fine‑tuning, yielding notable gains on question answering, fact verification, and arithmetic reasoning \cite{zhang2024chain,zhang2024learning}.  Beyond step selection, \citet{wu2025more} show that extremely long chains do not always improve performance and identify task‑specific optimal lengths; they advocate letting models decide when to “stop thinking’’ rather than imposing a fixed budget.  Complementary to this, \citet{li2025selfbudgeter} introduce \emph{SelfBudgeter}, which first predicts the minimal token allocation required for a given query and then generates a response that respects either the self‑estimated or a user‑defined budget, cutting decoding cost by 35–50\% with no accuracy loss. 

\paragraph{Curriculum Training}
Curriculum Training (CT) presents data in an easy‑to‑hard order to speed convergence and improve robustness \cite{bengio2009curriculum,hacohen2019power,zhang2024micm,pmlr-v235-zou24c}.  While CT works well in vision, NLP, and multimodal tasks, defining difficulty is challenging when complexity is semantic, perceptual, and relational.  FASTCURL \cite{song2025fastcurl} sidesteps manual heuristics by using prompt length as a proxy for reasoning complexity, gradually widening the context window.  This staged exposure reduces redundant reasoning and reaches state‑of‑the‑art accuracy on MATH 500 and OlympiadBench with fewer updates.
Building on these ideas, \citet{song2025fastcurl,zou2024a} extend curriculum reinforcement learning to reasoning models: they segment data by input length and progressively enlarge the context window during RLHF, achieving 49.6\% on AIME 2024 with a 1.5B‑parameter model.  Their ablation shows that both length‑aware segmentation and progressive window extension are necessary; removing either halves the gain. Apart from prompt length, other works estimate difficulty by model‑based uncertainty \cite{liu2024pacedd}, latent semantic novelty \cite{xu2024latent,Zou_2024_CVPR,zou2021annotation,zhang2024micm,xiaoevery,zeng2025objects,chen2025automated}, or retrieval distance \cite{chen2024retrieval}.  Adaptive schedulers that jointly tune pacing and difficulty metric—e.g., Dynamic Temperature Scaling (DTS) \cite{lee2025adaptive}—achieve faster convergence and better out‑of‑distribution robustness.  However, sub‑optimal curricula can still stall learning or trap models in local minima, underscoring the need for principled theory that couples data, capacity, and task structure.
\section{Role of CoT in Visual Grounding}

A \emph{Chain-of-Thought} (CoT) explicitly guides large language models (LLMs) through intermediate reasoning steps, which significantly improves their accuracy in arithmetic, commonsense reasoning, and symbolic reasoning tasks~\citep{wei2022chain,kojima2022large}. Recent vision-language models (VLMs) adapt CoT via prompt tuning or supervised fine-tuning methods, demonstrating effectiveness in multi-modal scenarios~\citep{ge2023chain,chen2024visual}. Beyond supervised methods, reinforcement learning (RL) provides another mechanism for enhancing reasoning capabilities. Specifically, \emph{Group Relative Policy Optimization} (GRPO)~\cite{shao2024deepseekmath} modifies standard Proximal Policy Optimization by employing variance-normalized relative advantage comparisons within generated response groups, removing the dependency on a learned critic, thereby achieving superior training stability and data efficiency~\citep{guo2025deepseek,shao2024deepseekmath}. GRPO has successfully enhanced reasoning performance in mathematical tasks~\citep{guo2025deepseek} and universal visual grounding~\citep{bai2025univg}, highlighting its potential for broader visual reasoning contexts.

Although CoT prompting generally benefits reasoning tasks, recent evidence suggests this improvement may not extend straightforwardly to visual grounding scenarios~\citep{wu2025more}. Specifically, longer reasoning chains may introduce excessive complexity or noisy reasoning steps, adversely impacting localization performance. To investigate this counterintuitive phenomenon, we systematically analyze how generating CoT affects visual grounding accuracy and explore whether the length of the CoT correlates with task difficulty.

\begin{figure}[!tbp]
\centering
\includegraphics[width=\linewidth]{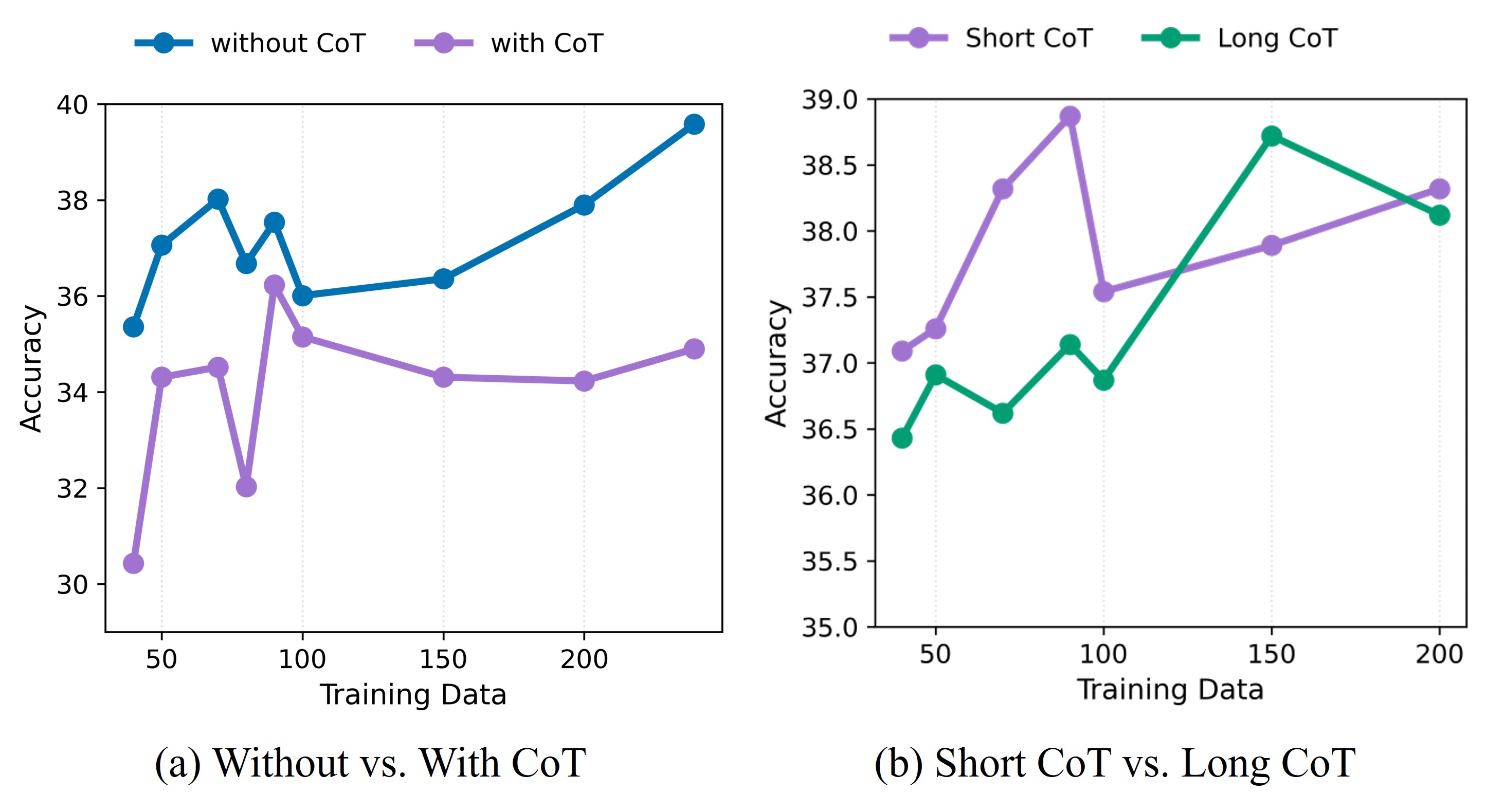}
\caption{(a) Accuracy comparison with and without CoT. (b) Performance comparison between short and long CoTs.}
\label{fig:a1}
\end{figure}

\subsection{Do CoTs Help Visual Grounding?}
\label{sec:effect_of_cot}

In our experiments, we fine-tune the visual grounding model using the GRPO algorithm guided by a reward function from~\cite{liu2025visual}. To evaluate the impact of explicitly outputting Chain-of-Thought (CoT) reasoning steps, we compare two experimental conditions (Table~\ref{tab:prompt}): (1) a scenario where the model directly outputs the final grounding results based on the given image and textual query (the "no CoT" condition), and (2) a scenario where the model explicitly generates intermediate CoT reasoning steps before producing the final prediction. As shown in Figure~\ref{fig:without_cot_better}, our findings reveal that the model consistently achieves superior performance when it is not required to explicitly output CoT. As shwon in Figure~\ref{fig:a1}(a), at a small dataset size of 40 samples, the model without CoT generation achieves an mIoU of 35.6, outperforming the CoT-outputting counterpart, which only reaches an mIoU of 34.3 even with a significantly larger dataset of 239 samples. Moreover, as the dataset expands, the performance gap widens further, with the direct-output model attaining a maximum mIoU of 39.6. These observations challenge the prevailing assumption that explicit intermediate reasoning steps invariably enhance model performance. Our empirical results suggest that, in visual grounding tasks, requiring explicit CoT generation can be unnecessary or even detrimental, particularly at smaller data scales, indicating potential inefficiencies or noise introduced by overly verbose reasoning chains.

\begin{figure}[!tbp]
\centering
\includegraphics[width=\linewidth]{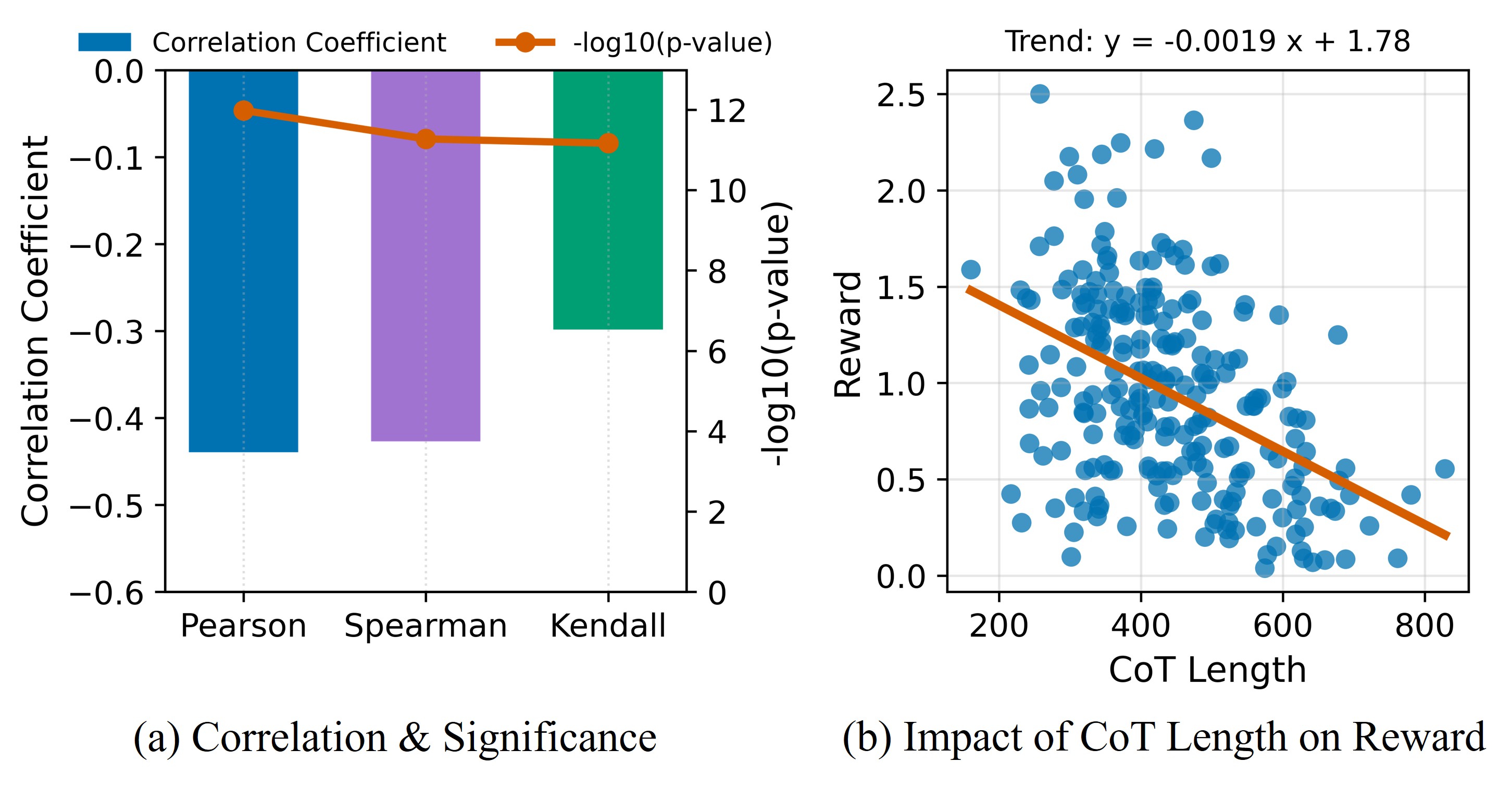}
\caption{(a) Correlation coefficients between CoT length and reward. (b) Negative impact of increased CoT length on reward.}
\label{fig:a2}
\end{figure}

\begin{figure*}[t]
\centering
\includegraphics[width=\linewidth]{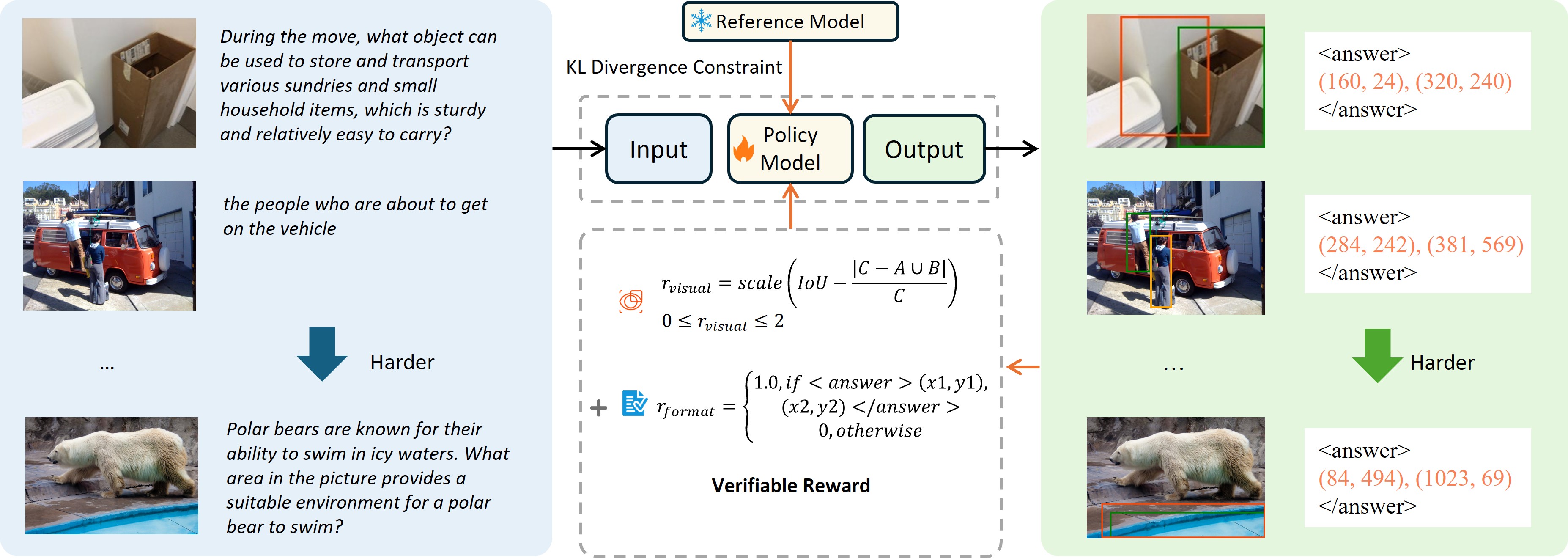}
\caption{Overview of curriculum-based GRPO training framework for visual grounding. We first sort training examples by the complexity indicated by their CoT length, from simplest (shortest CoT) to hardest (longest CoT). Each query-image pair is fed into a policy model constrained by KL-divergence with a reference model. The model outputs bounding boxes and receives a combined reward incorporating visual accuracy (scaled gIoU) and format correctness. This curriculum strategy progressively guides the model from simpler to increasingly complex visual grounding tasks.}
\label{GRPO}
\end{figure*}

\subsection{Does Longer CoT Mean Higher Difficulty?}

To understand whether CoT length significantly influences visual grounding performance, we empirically evaluate how variations in reasoning chain length correlate with model accuracy. Specifically, we generate multiple candidate CoTs for each visual grounding sample, compute their average length, and subsequently group samples into subsets of shorter and longer CoTs. Models trained on shorter-CoT subsets consistently achieve higher grounding accuracy (see Figure~\ref{fig:a1}(b)), suggesting that simpler reasoning chains are more beneficial during early learning stages. Moreover, models initially struggle with longer CoT samples—reflected by reduced accuracy—indicating increased task complexity associated with extended reasoning chains. Nonetheless, after continued training on larger datasets containing longer-CoT data, the model progressively improves and eventually surpasses performance obtained from shorter-CoT data. This finding clearly suggests that CoT length acts as an implicit indicator of task difficulty, directly impacting model learning dynamics and overall performance.

To theoretically underpin this empirical observation, we posit that the complexity of reasoning tasks increases with the number of required reasoning steps. Consider a simplified probabilistic model, where the probability of successfully completing a reasoning chain consisting of $C$ independent steps, each with success probability $p_c$, is given by: $\text{Pr}(\text{success}) = \prod_{c=1}^{C} p_c$. Assuming $p_c < 1$, increasing the number of reasoning steps $C$ (i.e., CoT length) exponentially decreases the probability of overall success, making longer reasoning chains inherently more challenging. Empirically, we verify this theoretical assumption by analyzing the correlation between CoT length and the generalized Intersection-over-Union (gIoU)-based reward. As shown in Figure~\ref{fig:a2}(a), our statistical analysis reveals a clear negative correlation between CoT length and reward: the Pearson correlation coefficient is $-0.4395$ ($p=1.04\times 10^{-12}$), Spearman's rank correlation coefficient is $-0.4268$ ($p=5.34\times 10^{-12}$), and Kendall's Tau coefficient is $-0.2981$ ($p=6.81\times 10^{-12}$). These results robustly confirm that increased CoT length systematically corresponds to decreased reward, reflecting higher task difficulty. Additionally, Figure~\ref{fig:a2}(b) visually confirms this inverse relationship: longer CoTs are systematically associated with lower rewards, reflecting increased difficulty. Taken together, these theoretical and empirical analyses affirm that CoT length serves as a robust signal of task complexity in visual grounding tasks.

\subsection{Does More Data Always Mean Better?}

Further investigation into the impact of training data scale on visual grounding performance reveals intriguing trends. As depicted in Figure~\ref{fig:a1}(a), simply increasing the size of the training dataset does not guarantee consistent performance improvements. Surprisingly, the model explicitly generating CoT demonstrates significant performance instability, with accuracy initially increasing but subsequently dropping and remaining stagnant even as more data is introduced. Conversely, the direct-output model without CoT consistently improves with increasing data size, highlighting the influence of output strategies on performance stability. Furthermore, Figure~\ref{fig:a1}(b) examines performance variations in terms of CoT length, indicating that models trained on shorter-CoT subsets initially exhibit higher accuracy compared to those trained on longer-CoT subsets. Nevertheless, with larger training datasets, the performance gap diminishes, suggesting that longer reasoning chains, despite their inherent complexity, become manageable as the model gradually adapts. These observations collectively emphasize the importance of data complexity and suggest that merely enlarging the dataset size is insufficient. Instead, careful consideration of data ordering and complexity progression may enhance the effectiveness of training visual grounding models.

\begin{table}[!tbp]
  \centering
  \caption{\textbf{Prompts used to construct the dataset.} Prompts are shown separately for direct output and CoT-based output.}
  \label{tab:prompt}
  \begin{tabularx}{\linewidth}{@{}lX@{}}
    \toprule
    \textbf{Direct prompt} &
    {Question} Output your grounding box. Following \lstinline|"<answer>(x1,y1),(x2,y2)</answer>"| format.\\[5pt]
    \midrule
    \textbf{CoT prompt} &
    {Question} Output the thinking process in \lstinline|"<think>...</think>"| and then the grounding box, following the format: \lstinline|"<think>reasoning chain</think><answer>(x1,y1),(x2,y2)</answer>"|.\\[5pt]
    \bottomrule
  \end{tabularx}
\end{table}

\section{Method}

Motivated by our key empirical observation that longer Chain-of-Thought (CoT) reasoning is associated with increased task difficulty and lower rewards, we propose a novel training framework termed Curriculum-based Relative Policy Optimization (CuRPO). CuRPO leverages a curriculum learning paradigm to progressively guide the model from simpler visual grounding tasks, characterized by shorter CoTs, to more complex tasks involving longer and more intricate reasoning processes. By systematically organizing and gradually introducing training examples based on CoT length and their corresponding reward signals, CuRPO effectively enhances the model’s reasoning capacity and localization accuracy, while maintaining training stability and efficiency.

\subsection{Reward Function Design}

In visual grounding tasks, a standard metric for assessing localization accuracy is Intersection over Union (IoU). However, IoU cannot provide useful feedback when predicted and ground-truth bounding boxes do not overlap, as the intersection area becomes zero. To resolve this limitation, we adopt Generalized IoU (gIoU)~\citep{rezatofighi2019generalized}, which considers both overlap and spatial relationships between bounding boxes. Formally, gIoU is defined as:
\begin{equation}
\text{gIoU}(A,B) = \text{IoU}(A,B) - \frac{C - (A \cup B)}{C},
\end{equation}
where \(C\) denotes the area of the smallest enclosing box containing predicted box \(A\) and ground-truth box \(B\). This allows meaningful feedback even without overlap.

We define our overall reward function as:
\begin{equation}
R_d = R_{\text{visual}} + R_{\text{format}},
\end{equation}
where \(R_{\text{visual}}\) is derived from the gIoU metric, and \(R_{\text{format}}\) ensures compliance with output formatting requirements. To stabilize training, we linearly rescale gIoU values from their original range \([-1,1]\) to \([0,2]\), reducing overly negative feedback and enhancing the visibility of spatial cues.

\subsection{Curriculum Training Process}

Given our observation that CoT length influences task difficulty, we propose a curriculum training strategy based on sorting examples by their CoT lengths. Specifically, we first instruct a pretrained VLM to generate multiple CoTs (typically 8 per sample) for each data point, explicitly capturing its reasoning process.

Next, we sort training examples by their average CoT length (shortest to longest). Shorter CoTs typically correspond to simpler reasoning tasks and thus provide a natural ordering of task complexity. We then sequentially introduce samples into training:
\begin{enumerate}
    \item Initial training phase: the model sees only samples with short CoTs, learning fundamental visual reasoning patterns.
    \item Intermediate phase: gradually introduce medium-length CoTs, exposing the model to moderately complex reasoning scenarios.
    \item Advanced training phase: incorporate longer CoTs, challenging the model to reason through increasingly complex visual grounding tasks.
\end{enumerate}
This progressive approach ensures the model acquires robust reasoning capabilities before tackling the most difficult examples.

\begin{algorithm}[!tbp]
\caption{Curriculum-based Relative Policy Optimization (CuRPO)}
\label{alg:CuRPO}
\small
\begin{algorithmic}[1]
\REQUIRE Dataset $\mathcal{D}$; initial policy $\pi_{\theta}$; reference policy $\pi_{\text{ref}}$; \textbf{SortCriterion} $\in\{\textit{Length},\textit{Reward}\}$; total training steps $T$; group size $G$; clip ratio $\epsilon$; KL weight $\beta$
\ENSURE Optimized policy $\pi_{\theta}$

\STATE Compute complexity score $s(x)$ for each sample $x\!\in\!\mathcal{D}$ according to \textbf{SortCriterion}  
\STATE Sort $\mathcal{D}$ in ascending order of $s(x)$ and split into curriculum phases $\{\mathcal{D}_1,\dots,\mathcal{D}_M\}$

\FOR{phase $m=1$ \TO $M$}
    \FOR{$t=1$ \TO $T/M$}
        \STATE Sample mini-batch $\mathcal{B}\subset\mathcal{D}_m$
        \FORALL{$(q,I)\in\mathcal{B}$}
            \STATE Generate $G$ candidate outputs $\{o_i\}_{i=1}^G\sim\pi_{\theta}(o\mid q,I)$
            \STATE Compute reward $r_i' = R_{\text{visual}}(o_i) + R_{\text{format}}(o_i)$
            \STATE Compute $\mu'=\tfrac{1}{G}\sum_{j=1}^G r_j'$, \;$\sigma'=\sqrt{\tfrac{1}{G}\sum_{j=1}^G(r_j'-\mu')^{2}}$
            \STATE $A_i \leftarrow \dfrac{r_i'-\mu'}{\sigma'}$  \COMMENT{group-normalized advantage}
            \STATE $c_i \leftarrow \dfrac{\pi_{\theta}(o_i\mid q,I)}{\pi_{\text{old}}(o_i\mid q,I)}$
            \STATE $L_i \leftarrow \min\!\bigl(c_iA_i,\;\text{clip}(c_i,1-\epsilon,1+\epsilon)A_i\bigr)$
        \ENDFOR
        \STATE $\displaystyle
               \theta \leftarrow \theta +
               \eta\nabla_{\theta}\!\Bigl(\frac{1}{|\mathcal{B}|G}\sum_{(q,I)\in\mathcal{B}}\sum_{i=1}^G L_i
               -\beta\,D_{\text{KL}}\bigl(\pi_{\theta}\,\|\,\pi_{\text{ref}}\bigr)\Bigr)$
    \ENDFOR
\ENDFOR
\RETURN $\pi_{\theta}$
\end{algorithmic}
\end{algorithm}

\subsection{GRPO-based Training Objective}

Inspired by the Visual-RFT~\citep{liu2025visual}, we employ Group Relative Policy Optimization (GRPO)~\citep{shao2024deepseekmath} as our reinforcement learning backbone. GRPO optimizes relative advantages within groups of generated responses, significantly enhancing sample efficiency and stability compared to standard PPO variants.

Given a query \(q\), the current policy \(\pi_{\theta}\) generates a group of \(G\) candidate outputs \(\{o_i\}\), each associated with a modified reward \(r_i'=R_d(o_i|q)\) that implicitly penalizes long CoTs through lower gIoU scores. We first compute group-normalized advantages:
\begin{equation}
A_i = \frac{r_i' - \mu'}{\sigma'}, \quad \mu'=\frac{1}{G}\sum_{j=1}^{G} r_j', \quad \sigma'=\sqrt{\frac{1}{G}\sum_{j=1}^{G}(r_j'-\mu')^2}.
\end{equation}

The final GRPO objective is defined as a clipped surrogate loss plus KL-divergence regularization to maintain policy stability:
\begin{align}
L_{\text{GRPO}}(\theta) &= -\frac{1}{G}\sum_{i=1}^{G}\min\bigl(c_i A_i,\;\text{clip}(c_i, 1-\epsilon, 1+\epsilon) A_i\bigr) \nonumber \\
&\quad - \beta\, D_{\text{KL}}\bigl(\pi_{\theta}(\cdot|q)\,\|\,\pi_{\text{ref}}(\cdot|q)\bigr),
\end{align}
where \(c_i=\frac{\pi_{\theta}(o_i|q)}{\pi_{\text{old}}(o_i|q)}\) represents the probability ratio between new and old policies, and \(\pi_{\text{ref}}\) serves as a stable reference policy.


Algorithm~\ref{alg:CuRPO} condenses the entire CuRPO pipeline.
We first assign each image–query pair a \emph{complexity score}, either the average CoT length or its corresponding reward, and sort the dataset from easiest to hardest before splitting it into curriculum phases.
Within every phase, the policy $\pi_{\theta}$ generates \(G\) candidate responses, receives a combined reward that blends scaled gIoU and format‐correctness, and converts these scores into \emph{group‐normalised advantages}.
CuRPO then maximises a clipped surrogate loss while regularising with a KL term that keeps the updated policy close to a reference model $\pi_{\text{ref}}$.
By gradually unlocking harder phases, the policy is forced to master short, easy reasoning chains first and then adapt to longer, more complex CoTs.

\section{Experiments}
\label{sec:exp}

\subsection{Experimental Setup}

\begin{table}[!tbp]
\centering
\caption{Results on LISA.  Visual grounding performance for different methods and training-data sizes.  
To save space, the citation for each baseline is placed on a second line within the same cell. Each experiment is repeated three times with different random seeds.}
\label{tab:dataset_lisa}
\setlength{\tabcolsep}{6pt}  
\begin{adjustbox}{max width=\linewidth}
\begin{tabular}{@{}lccc@{}}
\toprule
\textbf{Method} & \textbf{Model} & \textbf{\#Train} & \textbf{mIoU} \\
\midrule
\begin{tabular}[c]{@{}l@{}}SFT\\\cite{visual_rft}\end{tabular}                  & Qwen2-VL-2B & 239 & 29.7 \\
\begin{tabular}[c]{@{}l@{}}OV-Seg\\\cite{liang2023open}\end{tabular}             & OV-Seg      & 239 & 30.5 \\
\begin{tabular}[c]{@{}l@{}}GroundingDINO\\\cite{liu2024grounding}\end{tabular}   & X-Decoder   & 239 & 28.5 \\
\begin{tabular}[c]{@{}l@{}}GroundedSAM\\\cite{zou2023generalized}\end{tabular}   & X-Decoder   & 239 & 28.6 \\
\begin{tabular}[c]{@{}l@{}}Visual-RFT\\\cite{liu2025visual}\end{tabular}         & Qwen2-VL-2B & 239 & 34.4 \\
\midrule
\rowcolor{cyan!10}
\textbf{CuRPO (Ours)} & Qwen2-VL-2B &  50 & 37.4 {\color{red}{(+3.0)}} \\
\rowcolor{cyan!10}
\textbf{CuRPO (Ours)} & Qwen2-VL-2B & 200 & \textbf{38.7} {\color{red}{(+4.3)}} \\
\rowcolor{cyan!10}
\textbf{CuRPO (Ours)} & Qwen2-VL-2B & 239 & \textbf{38.4} {\color{red}{(+4.0)}} \\
\bottomrule
\end{tabular}
\end{adjustbox}
\end{table}

\paragraph{Implementation Details.}

Building on preliminary insights, we refine our curriculum training by using a more granular sorting strategy. Specifically, we sort training examples first by CoT length and then within each CoT-length bin (interval = 50 tokens) by reward values. This ensures that the model sees easiest examples (short CoTs and high reward) first, and gradually transitions to more complex ones. By progressively introducing longer CoTs and lower-reward samples, the curriculum training better scaffolds the model’s visual reasoning development. We conduct training using the Qwen2‑VL‑2B model~\citep{wang2024qwen2} fine‑tuned with GRPO. Our baseline is the zero‑curriculum fine‑tuned Qwen‑VL‑2B model in the “with CoT” setting: the model is required to generate a full Chain‑of‑Thought before outputting its final bounding box. We evaluate its mIoU and mAP performance to compare against our curriculum‑trained versions.

\paragraph{Datasets and Metrics.}
We conduct experiments on two standard vision‑language grounding datasets: RefCOCO and LISA.
\textbf{RefCOCO}~\cite{refcoco_dataset} is a widely used referring-expression comprehension benchmark, where the model must localize objects referenced by natural language.  
\textbf{LISA}~\cite{LISA_dataset} is a multi-scene visual grounding benchmark requiring robust localization under complex layouts. 
We report class-wise Average Precision (AP) and mean Average Precision (mAP) following standard evaluation protocols~\citep{padilla2020survey}. 

\subsection{Main Results}

\paragraph{Results on LISA.}
The experimental results on the LISA dataset are summarized in Table~\ref{tab:dataset_lisa}. Our proposed \textbf{CuRPO} consistently outperforms all baseline methods. Notably, with only 50 training examples, \textbf{CuRPO} achieves an mIoU of 37.4, exceeding the strongest baseline Visual-RFT~\cite{liu2025visual} trained on all 239 examples by +3.0 points. Increasing the training data to 200 examples further boosts performance to 38.7 mIoU, with gains remaining above +4.0 points over Visual-RFT when using all 239 training examples. These results support our hypothesis that curriculum learning based on CoT length substantially enhances visual grounding performance, especially in low-data regimes, with improvements that grow and then saturate as more training data becomes available.

\begin{table}[!tbp]
\centering
\caption{Comparison results (mAP) of different methods on RefCOCO, RefCOCO+ and RefCOCOg datasets. Each CuRPO experiment is repeated 30 times with six different random seeds.}
\label{tab:merged_results}
\begin{adjustbox}{max width=\linewidth}
\begin{tabular}{@{}l l c@{}}
\toprule
\textbf{Dataset} & \textbf{Method} & \textbf{mAP} \\
\midrule
\multirow{5}{*}{RefCOCO (val)} & Qwen2-VL-2B & 11.57 \\
& Visual-RFT~\cite{liu2025visual} & 21.28 \\
& \cellcolor{cyan!10}\textbf{CuRPO} (Random) & \cellcolor{cyan!10}33.09 (\textcolor{red}{+11.81}) \\
& \cellcolor{cyan!10}\textbf{CuRPO} (Length) & \cellcolor{cyan!10}\textbf{33.80 (\textcolor{red}{+12.52})} \\
& \cellcolor{cyan!10}\textbf{CuRPO} (Reward) & \cellcolor{cyan!10}32.64 (\textcolor{red}{+11.36}) \\
\midrule
\multirow{5}{*}{RefCOCO (test)} & Qwen2-VL-2B & 10.70 \\
& Visual-RFT~\cite{liu2025visual} & 20.38 \\
& \cellcolor{cyan!10}\textbf{CuRPO} (Random) & \cellcolor{cyan!10}29.92 (\textcolor{red}{+9.54}) \\
& \cellcolor{cyan!10}\textbf{CuRPO} (Length) & \cellcolor{cyan!10}\textbf{31.42 (\textcolor{red}{+11.04})} \\
& \cellcolor{cyan!10}\textbf{CuRPO} (Reward) & \cellcolor{cyan!10}27.89 (\textcolor{red}{+7.51}) \\
\midrule
\multirow{5}{*}{RefCOCO+ (val)} & Qwen2-VL-2B & 13.72 \\
& Visual-RFT~\cite{liu2025visual} & 18.41 \\
& \cellcolor{cyan!10}\textbf{CuRPO} (Random) & \cellcolor{cyan!10}26.82 (\textcolor{red}{+8.41}) \\
& \cellcolor{cyan!10}\textbf{CuRPO} (Length) & \cellcolor{cyan!10}26.18 (\textcolor{red}{+7.77}) \\
& \cellcolor{cyan!10}\textbf{CuRPO} (Reward) & \cellcolor{cyan!10}\textbf{26.85 (\textcolor{red}{+8.44})} \\
\midrule
\multirow{5}{*}{RefCOCO+ (test)} & Qwen2-VL-2B & 16.11 \\
& Visual-RFT~\cite{liu2025visual} & 20.90 \\
& \cellcolor{cyan!10}\textbf{CuRPO} (Random) & \cellcolor{cyan!10}24.34 (\textcolor{red}{+3.44}) \\
& \cellcolor{cyan!10}\textbf{CuRPO} (Length) & \cellcolor{cyan!10}\textbf{25.10 (\textcolor{red}{+4.20})} \\
& \cellcolor{cyan!10}\textbf{CuRPO} (Reward) & \cellcolor{cyan!10}23.55 (\textcolor{red}{+2.65}) \\
\midrule
\multirow{5}{*}{RefCOCOg (val)} & Qwen2-VL-2B & 14.89 \\
& Visual-RFT~\cite{liu2025visual} & 23.39 \\
& \cellcolor{cyan!10}\textbf{CuRPO} (Random) & \cellcolor{cyan!10}27.98 (\textcolor{red}{+4.59}) \\
& \cellcolor{cyan!10}\textbf{CuRPO} (Length) & \cellcolor{cyan!10}29.27 (\textcolor{red}{+5.88}) \\
& \cellcolor{cyan!10}\textbf{CuRPO} (Reward) & \cellcolor{cyan!10}\textbf{32.65 (\textcolor{red}{+9.26})} \\
\bottomrule
\end{tabular}
\end{adjustbox}
\label{tab:dataset_statistics}
\end{table}

\paragraph{Results on RefCOCO, RefCOCO+ and RefCOCOg.}
Table~\ref{tab:merged_results} summarizes the experimental results on RefCOCO, RefCOCO+, and RefCOCOg. Across all splits, our \textbf{CuRPO} variants consistently outperform the baseline models (Qwen2-VL-2B and Visual-RFT~\cite{liu2025visual}), highlighting the effectiveness of curriculum-based training.
For the RefCOCO dataset, \textbf{CuRPO (Length)} achieves the highest mAP of 33.80 and 31.42 on the validation and test sets, respectively, surpassing Visual-RFT by {+12.52} and {+11.04}. Even \textbf{CuRPO (Random)} already provides large gains (33.09/29.92), indicating that curriculum RL itself is highly beneficial, while ordering examples by CoT length further improves performance.
On the RefCOCO+ dataset, all CuRPO variants bring clear improvements over Visual-RFT. \textbf{CuRPO (Reward)} attains the best validation mAP of 26.85 ({+8.44}), whereas \textbf{CuRPO (Length)} yields the best test mAP of 25.10 ({+4.20}). The close performance of length-based and reward-based curricula suggests that structured difficulty signals become increasingly useful as textual descriptions grow more complex.
For RefCOCOg, which contains longer and more ambiguous descriptions, \textbf{CuRPO (Reward)} achieves the highest mAP of 32.65, a substantial {+9.26} improvement over Visual-RFT, while the length-based curriculum also performs strongly. These results collectively confirm that curriculum learning based on CoT length and reward signals substantially boosts visual grounding performance across datasets of varying difficulty.

\subsection{Ablation studies}

\begin{figure}[!tbp]
\centering
\includegraphics[width=\linewidth]{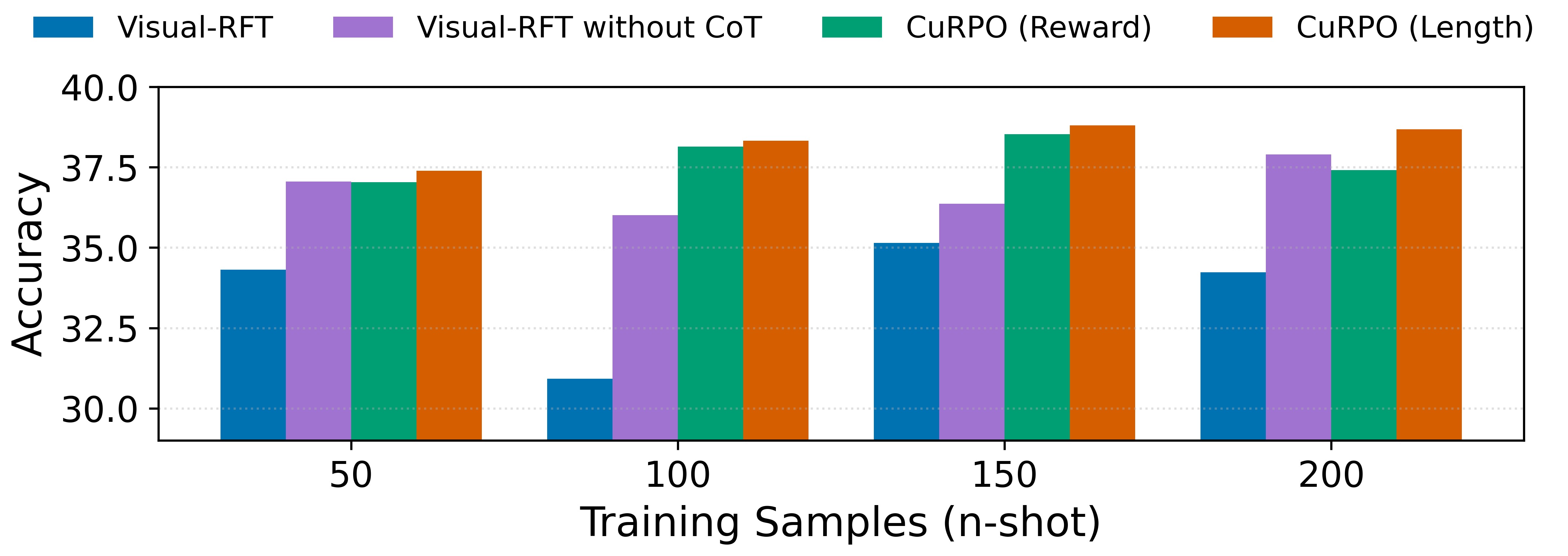}
\caption{(Comparison of visual grounding accuracy across different training sample sizes and sorting methods.}
\label{fig:a3}
\end{figure}

\paragraph{Impact of Sorting Strategies.} 
As shown in Figure~\ref{fig:a3} and Table~\ref{tab:merged_results}, all three sorting strategies yield substantial gains over the non-curriculum baseline. On RefCOCO, \textbf{CuRPO (Length)} consistently outperforms both \textbf{CuRPO (Random)} and \textbf{CuRPO (Reward)}, indicating that ordering samples by CoT length is particularly effective on this relatively simple dataset. For RefCOCO+, \textbf{CuRPO (Reward)} achieves the best validation mAP, while \textbf{CuRPO (Length)} performs best on the test split, showing that both length- and reward-based curricula provide strong and complementary benefits. On the more challenging RefCOCOg dataset, \textbf{CuRPO (Reward)} clearly dominates, suggesting that directly exploiting reward signals becomes more important when descriptions are long and ambiguous. Notably, even \textbf{CuRPO (Random)}—which does not explicitly encode task difficulty—still brings large improvements over the baseline, implying that curriculum RL without explicit CoT generation is already beneficial. Overall, leveraging CoT length and reward signals to structure training provides robust gains, with reward-based ordering particularly advantageous on the most difficult dataset.

\paragraph{Effect of CoT across Data Scales.}
In Figure~\ref{fig:a3}, we further investigate model performance across different numbers of training samples. Interestingly, results indicate that the model explicitly generating CoTs (Visual-RFT) consistently underperforms the same model trained without explicit CoT generation. This aligns with our earlier hypothesis that explicit CoT generation in visual grounding introduces unnecessary complexity, potentially confusing the model and negatively affecting localization accuracy. In contrast, models trained without explicit CoT generation consistently achieve better accuracy, verifying that direct learning reduces reasoning ambiguity, thereby simplifying task acquisition. Moreover, Figure~\ref{fig:a3} demonstrates that CuRPO-Length consistently outperforms both baseline conditions (Visual-RFT with and without CoT) across varying amounts of data. Notably, CuRPO methods show steady improvements as the dataset size increases, highlighting the advantages of curriculum-based approaches. This suggests that structured task complexity progression significantly enhances model generalization, especially as more data becomes available.

\paragraph{Superior Few-Shot Performance.}
Our analysis emphasizes CuRPO's remarkable effectiveness in few-shot scenarios (50 samples). At this scale, CuRPO methods substantially outperform baselines, demonstrating that explicitly removing the requirement for CoT generation allows the model to rapidly learn visual reasoning patterns without unnecessary intermediate reasoning complexity. The rapid performance gain observed in few-shot conditions underscores CuRPO's capacity for efficiently leveraging limited data. Thus, our proposed framework proves particularly advantageous when faced with limited data resources, quickly capturing essential grounding capabilities by reducing the cognitive overhead introduced by unnecessary reasoning steps.

\section{Conclusion}
\label{sec:con}
In this paper, we have systematically investigated the role of CoT prompting in visual grounding tasks, revealing that explicitly generating CoT does not universally benefit model performance and may even hinder accuracy, especially with longer reasoning chains. Motivated by our empirical analysis, we introduced CuRPO, a reinforcement learning method that leverages CoT length and reward signals to progressively structure the training complexity. CuRPO effectively guides the model from simpler reasoning tasks to increasingly challenging ones, resulting in substantial improvements in localization accuracy, enhanced reasoning capabilities, and stable training behavior even under few-shot conditions. Extensive experiments across diverse visual grounding benchmarks validate that our curriculum-driven strategy consistently outperforms existing approaches, demonstrating the importance of carefully ordered training data for optimizing reasoning-intensive vision-language tasks. Future work will focus on evaluating and adapting this curriculum–GRPO paradigm to other vision–language and multimodal reasoning tasks to assess its generalization capacity beyond visual grounding.

\section{Acknowledgements}
This work is supported by the National Natural Science Foundation of China under grants 62206102, 62436003, 62376103, 62302184, and 62402015; the National Key Research and Development Program of China under grant 2024YFC3307900; the Major Science and Technology Project of Hubei Province under grants 2025BAB011 and 2024BAA008; the Hubei Science and Technology Talent Service Project under grant 2024DJC078; Ant Group through the CCF–Ant Research Fund; the Postdoctoral Fellowship Program of the China Postdoctoral Science Foundation under grant GZB20230024; and the China Postdoctoral Science Foundation under grant 2024M750100. Computations were performed on the HPC Platform of Huazhong University of Science and Technology.

\bibliography{reference}

\newpage
\appendix

\section{Per-Category Localization Performance}

To better understand how different curriculum strategies affect localization across object categories, we report per-category mean Average Precision (mAP) for RefCOCO, RefCOCO+, and RefCOCOg. Each table compares the pretrained baseline, direct CoT supervision, and three curriculum strategies (random, CoT-length based, and reward-based ordering). These results highlight that curriculum-based training not only improves overall mAP, but also yields consistent gains on challenging categories such as \emph{chair}, \emph{bed}, and \emph{toilet}, where precise grounding is more difficult.

\begin{table}[!htbp]
\centering
\caption{Per-category mAP on the RefCOCO validation split.}
\begin{adjustbox}{max width=\linewidth}
\begin{tabular}{lccccccccc}
\hline
Method & mAP & bus & train & chair & airplane & cat & dog & bed & toilet \\
\hline
pretrained & 11.57 & 15.90 & 8.49 & 1.89 & 13.31 & 15.84 & 21.53 & 4.73 & 10.86 \\
with cot & 21.28 & 22.65 & 22.82 & 14.18 & 12.86 & 21.57 & 34.87 & 12.23 & 29.07 \\
random & 33.09 & 43.68 & 25.85 & 25.07 & 17.94 & 40.05 & 50.49 & 20.17 & 41.44 \\
cot length & 33.80 & 44.17 & 27.09 & 25.69 & 19.39 & 41.84 & 51.75 & 20.95 & 39.55 \\
reward & 32.64 & 41.22 & 20.85 & 23.09 & 15.58 & 43.80 & 53.27 & 18.27 & 45.05 \\
\hline
\end{tabular}
\end{adjustbox}

\bigskip
\centering
\caption{Per-category mAP on the RefCOCO test split.}
\begin{adjustbox}{max width=\linewidth}
\begin{tabular}{lccccccccc}
\hline
Method & mAP & bus & train & chair & airplane & cat & dog & bed & toilet \\
\hline
pretrained & 10.70 & 25.17 & 9.91 & 5.48 & 6.33 & 8.58 & 9.23 & 8.64 & 12.25 \\
with cot & 20.38 & 38.60 & 23.70 & 12.68 & 7.58 & 28.31 & 23.88 & 8.64 & 19.67 \\
random & 29.92 & 48.17 & 32.66 & 21.79 & 13.55 & 47.46 & 39.03 & 9.31 & 27.44 \\
cot length & 31.42 & 50.96 & 34.00 & 22.44 & 15.90 & 49.70 & 39.55 & 9.74 & 29.04 \\
reward & 27.89 & 42.88 & 32.04 & 22.79 & 9.12 & 52.85 & 32.40 & 6.79 & 24.28 \\
\hline
\end{tabular}
\end{adjustbox}

\bigskip
\centering
\caption{Per-category mAP on the RefCOCO+ validation split.}
\begin{adjustbox}{max width=\linewidth}
\begin{tabular}{lccccccccc}
\hline
Method & mAP & bus & train & chair & airplane & cat & dog & bed & toilet \\
\hline
pretrained & 13.72 & 22.88 & 12.81 & 5.12 & 7.42 & 21.83 & 12.05 & 13.15 & 14.47 \\
with cot & 18.41 & 19.57 & 22.77 & 10.69 & 13.19 & 39.21 & 18.19 & 14.31 & 9.32 \\
random & 26.82 & 39.75 & 28.77 & 14.66 & 15.26 & 42.16 & 27.11 & 24.54 & 22.34 \\
cot length & 26.18 & 29.39 & 30.84 & 12.60 & 11.30 & 42.87 & 32.09 & 23.96 & 26.36 \\
reward & 26.85 & 40.57 & 28.59 & 15.48 & 18.98 & 42.06 & 23.33 & 18.82 & 26.96 \\
\hline
\end{tabular}
\end{adjustbox}

\bigskip
\centering
\caption{Per-category mAP on the RefCOCO+ test split.}
\begin{adjustbox}{max width=\linewidth}
\begin{tabular}{lccccccccc}
\hline
Method & mAP & bus & train & chair & airplane & cat & dog & bed & toilet \\
\hline
pretrained & 16.11 & 20.44 & 14.66 & 11.95 & 8.47 & 28.87 & 15.94 & 14.08 & 14.51 \\
with cot & 20.90 & 21.21 & 17.73 & 14.39 & 21.91 & 33.68 & 24.32 & 17.48 & 16.46 \\
random & 24.34 & 25.44 & 31.46 & 16.58 & 12.74 & 40.64 & 24.72 & 21.96 & 21.21 \\
cot length & 25.10 & 27.99 & 29.55 & 14.20 & 13.92 & 44.44 & 30.40 & 22.79 & 17.51 \\
reward & 23.55 & 28.42 & 19.77 & 18.84 & 19.06 & 34.81 & 22.82 & 22.28 & 22.38 \\
\hline
\end{tabular}
\end{adjustbox}

\bigskip
\centering
\caption{Per-category mAP on the RefCOCOg validation split.}
\begin{adjustbox}{max width=\linewidth}
\begin{tabular}{lccccccccc}
\hline
Method & mAP & bus & train & chair & airplane & cat & dog & bed & toilet \\
\hline
pretrained & 14.89 & 20.19 & 22.34 & 4.69 & 27.90 & 11.19 & 11.83 & 10.47 & 10.50 \\
with cot & 23.39 & 41.29 & 28.83 & 11.93 & 23.64 & 26.20 & 25.78 & 17.11 & 12.36 \\
random & 27.98 & 42.21 & 26.42 & 15.31 & 28.25 & 34.97 & 33.69 & 23.12 & 19.89 \\
cot length & 29.27 & 45.21 & 29.31 & 20.13 & 26.07 & 39.66 & 34.10 & 25.48 & 14.18 \\
reward & 32.65 & 44.72 & 36.00 & 18.93 & 28.05 & 35.39 & 45.18 & 29.64 & 23.33 \\
\hline
\end{tabular}
\end{adjustbox}
\end{table}

\section{Chain-of-Thought Quality and Usage}

\subsubsection{Empirical Analysis of CoT Quality.}

\begin{table*}[htbp]
\centering
\caption{Analysis of CoT quality and effectiveness based on similarity and reward thresholds. High-similarity CoTs align better with the ground-truth region, while high reward indicates successful localization.}
\begin{adjustbox}{max width=\linewidth}
\begin{tabular}{ccccc}
\hline
Image-Text Similarity & Reward & Count & Ratio (n / (239*8)) & Ratio (n / n\_similarity) \\
\hline
High similarity & High reward & 100 & 5.23 & 13.46 \\
High similarity & Low reward & 643 & 33.63 & 86.54 \\
Low similarity & High reward & 31 & 1.62 & 2.65 \\
Low similarity & Low reward & 1138 & 59.52 & 97.35 \\
\hline
\end{tabular}
\end{adjustbox}
\end{table*}

We compute two similarity scores for each sample: (1) the similarity between the model-generated chain-of-thought and the textual description of the ground-truth bounding box, and (2) the similarity between the input question and the same bounding-box description. A CoT is labeled \textbf{High similarity} if its CoT--bbox similarity exceeds the question--bbox similarity; otherwise, it is labeled \textbf{Low similarity}. For the localization reward (0~$\leq$~reward~$\leq$~3), we similarly categorize each instance into \textbf{High} or \textbf{Low} using the threshold 1.5.

The statistics in the table show that nearly 60\% of the CoTs fall into the low-similarity and low-reward region (59.52\%), and almost all low-quality CoTs fail to produce correct predictions (97.35\% within the low-similarity group). In contrast, high-quality CoTs are more likely to yield accurate localization (13.46\% of the high-similarity group), but the absolute fraction of such effective CoTs remains small (5.23\% of all samples). These observations motivate our curriculum design, which aims to prioritize informative, well-aligned CoTs while down-weighting noisy or misleading reasoning traces.

\subsubsection{Capacity Competition Between Reasoning and Localization.}

\begin{figure}[!tbp]
\centering
\includegraphics[width=\linewidth]{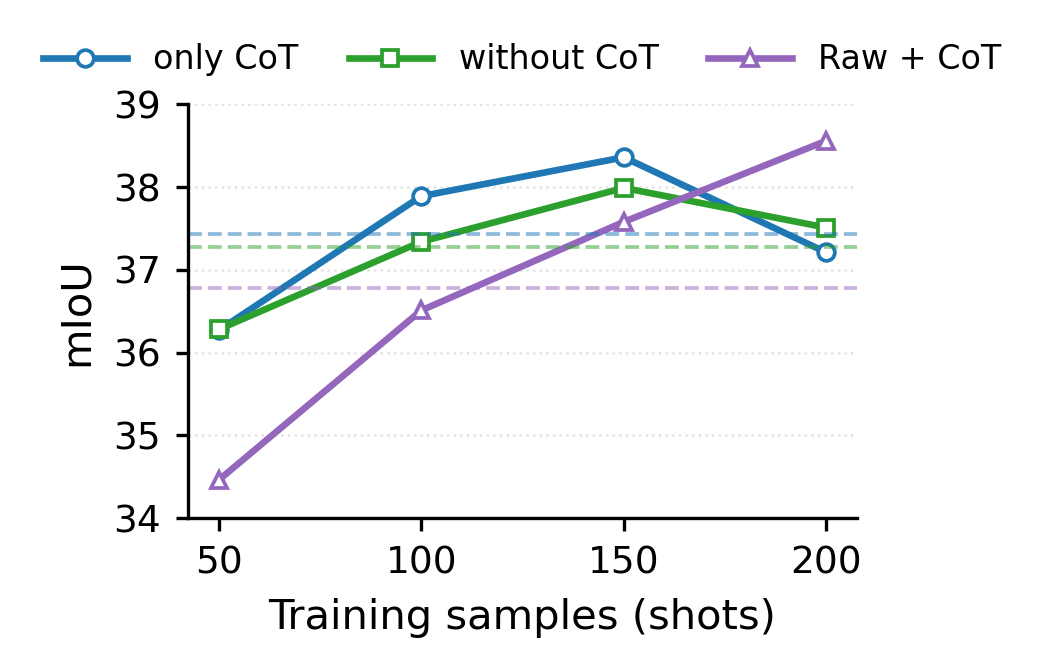}
\caption{Comparison of three text input settings: original question only, CoT-only, and Raw+CoT.}
\label{fig:cot as prompt}
\end{figure}

We hypothesize that jointly generating chain-of-thought explanations and predicting bounding boxes can lead to competition for model capacity, which may impair localization performance. To test this hypothesis, we evaluate three text input settings: (1) the original question only, (2) CoT-only input, and (3) a Raw+CoT configuration where the original question and its CoT are concatenated as a single prompt while the model focuses on predicting the bounding box.

At 200 training samples, Raw+CoT achieves an mIoU of 38.56, outperforming both the original-text setting (37.51) and the CoT-only setting (37.21). This result supports our capacity competition hypothesis and indicates that treating CoT as an auxiliary prompt, rather than a sequence to be generated in parallel with localization, allows the model to better allocate its representational capacity and improves grounding accuracy.

\section{Additional Results}

In this section, we supplment more results with the \textbf{Qwen2-VL-7B-Instruct} model to isolate the effect of a training strategies rather than architecture. Unless otherwise stated, we fix the random seed and train all models on the same hardware configuration, ensuring that improvements cannot be attributed to randomness or compute differences. The varying factors are the presence or absence of CoT supervision and the choice of curriculum strategy, as summarized in Table~\ref{tab:qwen2_performance}. Overall, we observe that introducing CoT without curriculum severely degrades performance compared with the pretrained model, while curriculum-based methods not only recover the loss but also bring substantial gains in mIoU. Among them, the CoT-length based curriculum achieves the best trade-off between training time and accuracy, suggesting that ordering samples from short to long reasoning chains provides a stable and effective learning trajectory for visual grounding.

\begin{table}[t]
\centering
\caption{Performance comparison under different training strategies using Qwen2-VL-7B-Instruct. All runs share the same seed and hardware.}
\label{tab:qwen2_performance}
\begin{adjustbox}{max width=\linewidth}
\begin{tabular}{lcccc}
\hline
Method & data size & Training Time & Val Time & mIoU \\
\hline
w/o cot & --      & --        & 04:54 & 39.45 \\
w cot    & --      & --        & 07:12 & 20.30 \\
Visual-RFT               & 239*6   & 2:23:21   & 08:38 & 45.29 \\
\midrule
\rowcolor{cyan!10}
\textbf{CuRPO}(Random)                 & 239*6   & 1:13:00   & 03:12 & \color{red}{51.21 (+5.92)} \\
\rowcolor{cyan!10}
\textbf{CuRPO}(Reward)               & 239*6   & 1:10:12   & 03:11 & \color{red}{52.08 (+6.79)} \\
\rowcolor{cyan!10}
\textbf{CuRPO}(Length)               & 239*6   & 1:11:55   & 03:12 & \textbf{\color{red}{52.66 (+7.37)}} \\
\hline
\end{tabular}
\end{adjustbox}
\end{table}

\end{document}